\documentclass[11pt,a4paper]{article}
\usepackage{acl2019}

\aclfinalcopy % Uncomment this line for the final submission
\def\aclpaperid{1669} %  Enter the acl Paper ID here
\usepackage{natbib}
\setcitestyle{square, comma, numbers,sort&compress}
\usepackage{times}
\usepackage{latexsym}
\usepackage{url}
\usepackage{amsmath}
\usepackage{graphicx}
\usepackage{caption}
\usepackage{subcaption}

\newcommand{\reffig}[1]{Figure~\ref{#1}}
\newcommand{\reftbl}[1]{Table~\ref{#1}}
\newcommand{\refsec}[1]{Section~\ref{#1}}

\newcommand{\reminder}[1]{}
\newcommand{\reply}[1]{}

\newcommand{\krnstwo}{PassAct2 dataset} 
\newcommand{\krnsfive}{Act3 dataset} 
\newcommand{\passacttwo}{PassAct1 dataset} 
\newcommand{\shortkrnstwo}{PassAct2} 
\newcommand{\shortkrnsfive}{Act3} 
\newcommand{\shortpassacttwo}{PassAct1} 

\newcommand*{\affaddr}[1]{#1} % No op here. Customize it for different styles.
\newcommand*{\affmark}[1][*]{\textsuperscript{#1}}
\newcommand*{\email}[1]{\texttt{#1}}

\newcommand*{\corpusname}[0]{Simple Sentence Corpus}

\title{Relating Simple Sentence Representations in Deep Neural Networks and the Brain }

\author{%
Sharmistha Jat\affmark[1]\thanks{\quad This research was carried out during a research internship
at the Carnegie Mellon University.} \quad
Hao Tang\affmark[2] \quad
Partha Talukdar\affmark[1] \quad 
Tom Mitchell\affmark[2] \\
\affaddr{\affmark[1]Indian Institute of Science, Bangalore}\\
\affaddr{\affmark[2]School of Computer Science, Carnegie Mellon University}\\
\email{
\{sharmisthaj,ppt\}@iisc.ac.in}\\
\email{
htang1@alumni.cmu.edu,
tom.mitchell@cs.cmu.edu}\\
}

\date{}

\begin{document}
\maketitle
\begin{abstract}

What is the relationship between sentence representations learned by deep recurrent models against those encoded by the brain? Is there any correspondence between hidden layers of these recurrent models and brain regions when processing sentences? Can these deep models be used to synthesize brain data which can then be utilized in other extrinsic tasks? We investigate these questions using sentences with simple syntax and semantics (e.g., \textit{The bone was eaten by the dog.}). We consider multiple neural network architectures, including recently proposed ELMo and BERT. We use magnetoencephalography (MEG) brain recording data collected from human subjects when they were reading these simple sentences. 

Overall, we find that BERT's activations correlate the best with MEG brain data. We also find that the deep network representation can be used to generate brain data from new sentences to augment existing brain data. 
To the best of our knowledge, this is the first work showing that the MEG brain recording when reading a word in a sentence can be used to distinguish earlier words in the sentence. Our exploration is also the first to use deep neural network representations to generate synthetic brain data and to show that it helps in improving subsequent stimuli decoding task accuracy.

\end{abstract}

\section{Introduction}

Deep learning methods for natural language processing have been very successful in a variety of Natural Language Processing (NLP) tasks. However, the representation of language learned by such methods is still opaque. The human brain is an excellent language processing engine, and the brain representation of language is of course very effective. Even though both brain and deep learning methods are representing language, the relationships among these representations are not thoroughly studied. \citet{WehbeEMNLP14} and \citet{hale_acl18} studied this question in some limited capacity. \citet{WehbeEMNLP14} studied the processing of a story context at a word level during language model computation. \citet{hale_acl18} studied the syntactic composition in  RNNG model \cite{dyer-rnng:16} with human encephalography (EEG) data.

We extend this line of research by investigating the following three questions: (1) what is the relationship between sentence representations learned by deep learning networks and those encoded by the brain; (2) is there any correspondence between hidden layer activations in these deep models and brain regions; and (3) is it possible for deep recurrent models to synthesize brain data so that they can effectively be used for brain data augmentation. In order to evaluate these questions, we focus on representations of simple sentences. %\reminder{TI : In order to evaluate these questions, we focused on representations of simple sentences. Tense keeps switching in this paragraph between past and present.}
We employ various deep network architectures, including recently proposed ELMo \cite{elmo2018} and BERT \cite{bert2018} networks. We use MagnetoEncephaloGraphy (MEG) brain recording data of simple sentences as the target  reference. We then correlate the representations learned by these various networks with the MEG recordings. Overall, we observe that BERT representations are the most predictive of MEG data. We also observe that the deep network models are effective at synthesizing brain data which are useful in overcoming data sparsity in stimuli decoding tasks involving brain data.

In summary, in this paper we make the following contributions.

\begin{itemize}
    \item We initiate a study to relate representations of simple sentences learned by various deep networks with those encoded in the brain. We establish correspondences between activations in deep network layers with brain areas.
    \item We demonstrate that deep networks are capable of predicting change in brain activity due to differences in previously processed words in the sentence.
    \item We demonstrate effectiveness of using deep networks to synthesize brain data for downstream data augmentation.
\end{itemize}

We have made our code and data\footnote{\url{https://github.com/SharmisthaJat/ACL2019-SimpleSentenceRepr-DNN-Brain}} publicly available to support further research in this area.

\section{Datasets}

In this section, we describe the MEG dataset and \corpusname{} used in the paper. 

\subsection{MEG Dataset}
\label{sec:megdataset}

%We report results on multiple simple sentence MEG datasets collected, summarised in \reftbl{tab:megdataset} \cite{nicoledata}. 
Magnetoencephalography (MEG) is a non-invasive functional brain imaging technique which records magnetic fields produced by electrical currents in the brain. Sensors in the MEG helmet allow for recording of magnetic fluctuations caused by changes in neural activity of the brain. For the experiments in this paper, we used three different MEG datasets collected when subjects were shown simple sentences as stimulus. These datasets are summarized in  \reftbl{tab:megdataset}, please see \cite{nicoledata} for more details. Additional dataset details are mentioned in appendix section \ref{supp:dataset3}. In the MEG helmet, 306 sensors were distributed over 102 locations and sampled at 1kHz. Native English speaking subjects were asked to read simple sentences. Each word within a sentence was presented for 300ms with 200ms subsequent rest. To reduce noise in the brain recordings, we represent a word's brain activity by averaging 10 sentence repetitions \cite{sudre2012}. Comprehension questions followed 10\% of sentences,
to ensure semantic engagement. MEG data was acquired using a 306 channel  Elekta Neuromag device. Preprocessing included spatial filtering using temporal signal space separation (tSSS), low-pass filtering  150Hz with notch filters at 60 and 120Hz, and downsampling to 500Hz \cite{WehbeEMNLP14}. Artifacts from tSSS-filtered same-day empty
room measurements, ocular and cardiac artifacts were removed via Signal Space Projection (SSP). 
%check the preprocessing parameters again.
    
   \begin{table}[]
\begin{tabular}{|c|c|c|c|}\hline
Dataset                          & \#Sentences & Voice & Repetition \\\hline
\shortpassacttwo{}   & 32          & P+A   & 10         \\\hline
\shortkrnstwo{}      & 32          & P+A   & 10         \\\hline
%\krnsthree{}    & 60          & A     & 10         \\\hline
%\krnsfour{}     & 60          & A     & 10         \\\hline
\shortkrnsfive{}     & 120         & A     & 10         \\\hline
%\passactthree{} & 32          & P+A   & 10        \\\hline
\end{tabular}
\caption{\label{tab:megdataset} MEG datasets used in this paper. Column `Voice' refers to the sentence voice, `P' is for passive sentences and `A' is for active. Repetition is the number of times the human subject saw a sentence. For our experiments, we average MEG data corresponding to multiple repetitions of a single sentence.}
\end{table}

\begin{figure*}[t]
\centering
  \setlength{\textfloatsep}{10.0pt}
  \includegraphics[scale=0.5]{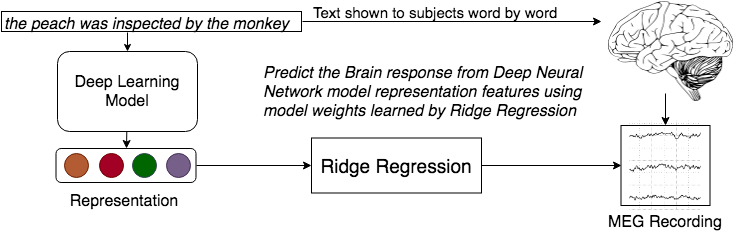}
  \caption{\label{fig:Encoding_Archi} Encoding model for MEG data. 306 channel 500ms MEG signal for a single word was compressed to $306 \times 5$ by averaging $100\textrm{ms}$ data into a single column. This MEG brain recording data is then encoded from text representation vector to brain activity using ridge regression. The evaluation is done using 5 fold cross-validation. Please see \refsec{sec:exp_res} for more details.}
\end{figure*}

\subsection{\corpusname{}}
    \label{sec:langdataset}
    In this paper, we aim to understand simple sentence processing in deep neural networks (DNN) and the brain. In order to train DNNs to represent simple sentences, we need a sizeable corpus of simple sentences. While the MEG datasets described in Section \ref{sec:megdataset} contain a few simple sentences, that set is too small to train DNNs effectively. In order to address this, we created a new \corpusname{} (SSC), consisting of a mix of simple active and passive sentences of the form
    ``\textit{the woman encouraged the girl"} and \textit{``the woman was encouraged by the boy"}, respectively. The SSC dataset consists of 256,145 sentences constructed using the following two sets.
    
    \begin{itemize}
        \item Wikipedia: We processed the 2009 Wikipedia dataset to get sentences  matching the following patterns. \\
         \textit{``the [noun+] was [verb+] by the [noun+]"}\\
         \textit{``the [noun+] [verb+] the [noun+]"}\\
        If the last word in the pattern matched is not noun, then we retain the additional dependent clause in the sentence. We were able to extract 117,690 active, and 8210 passive sentences from wikipedia.  
        \item NELL triples: In order to ensure broader coverage of Subject-Verb-Object (SVO) triples in our sentence corpus, we used the NELL SVO triples\footnote{NELL SVO triples: \url{http://rtw.ml.cmu.edu/resources/svo/}} \cite{Talukdar:2012:cikm}. %The NELL knowledge base is constructed using artificial intelligence, and therefore is not 100 \% correct. 
        We subsample SVO triples based on their frequency (threshold = 6), a frequent verb list, and Freebase to get meaningful sentences. Any triple with subject or object or verb not in Freebase is discarded from the triple set. %\reminder{PPT: describe more these filtering criteria}\reply{SJ: added one more line}We construct active and passive sentences as follows:
        \begin{itemize}
            \item Active sentence:  Convert the verb to its past tense and concatenate the triple using the following pattern: \textit{``the [subject] [verb-past-tense] the [object]"}.
            \item Passive sentence: Concatenate the triple using pattern: \textit{``the [object] was [verb-past-tense] by the [subject]"}
        \end{itemize}
        We generate 86,452 active and 43,793 passive sentences in total from the NELL triples. 
    \end{itemize}

    We train our deep neural network models with 90\% of sentences in this dataset and test on the remaining 10\%. We used the spaCy \cite{spacy2} library to predict POS tags for words in this dataset.

\section{Methods}

We test correlations between brain activity and deep learning model activations \cite{nature_deeplearning} for a given sentence using a classification task, similar to previous works \cite{2008Mitchell,WehbeStoryTrick2014,WehbeEMNLP14}. If we are able to predict brain activity from the neural network activation, then we hypothesize that there exists a relationship between the process captured by the neural network layer and the brain. The schematic of our encoding approach is shown in \reffig{fig:Encoding_Archi}.

We investigate various deep neural network models using context sensitivity tests to evaluate their performance in predicting brain activity. Working with these models and their respective training assumptions help us in understanding which assumption contributes to the correlations with the brain activity data. We process the sentences incrementally for each model to prevent information from future words from affecting the current representation, in line with how information is processed by the brain. For example, in the sentence \textit{``the dog ate the biscuit"}, the representation of the word \textit{``ate"} is calculated by processing sentence segment \textit{``the dog ate"} and taking the last representation in each layer as the context for the word \textit{``ate"}. The following embedding models are used to represent sentences. %to process sentences using neural networks.  

\begin{figure}[t]
\centering
  \setlength{\textfloatsep}{10.0pt}
  \includegraphics[scale=0.4]{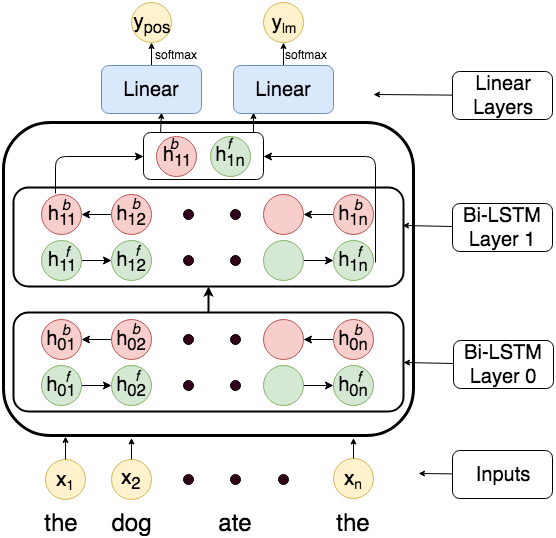}
  \caption{\label{fig:Multitask_archi} Architecture diagram for the simple multi-task model. The second LSTM layer's output is processed by 2 linear layers each producing the next-word and the POS-tag prediction. We process each sentence incrementally to get the prediction for word at the nth position, this helps in removing forward bias from future words and therefore is consistent with the information our brain receives when processing the same sentence. Our Simple Bi-directional LSTM language model also has a similar architecture with just one output linear layer for next word prediction.%\reminder{PPT: why show this fig at all? It is not central to our explorations and by now ppl should be quite familiar with such multitask archs. So, we may also skip the fig.} \reply{SJ: the description skips information about the architecture completely, which may be useful in understanding the results. We could remove this and add more description. Or, can we keep it if there is space ?}
  }
\end{figure}

\begin{itemize}
    \item \textbf{Random Embedding Model}: In this model, we represent each word in a context by a randomly generated 300-dimensional vector. Each dimension is uniformly sampled between [0,1]. The results from this model help us establish the random baseline.  
    
    \item \textbf{GloVe Additive Embedding Model}: This model represents a word context as the average of the current word's GloVe embedding \cite{pennington2014glove} and the previous word context. The first word in a sentence is initialized with its GloVe embedding as context. %This simple model's performance in tasks helps us contrast with the advanced model's representations. 
    
    \item \textbf{Simple Bi-directional LSTM Language Model}: We build a language model following \cite{DBLP:journals/corr/InanKS16}. Given a sequence of words $w_1 \ldots w_t$, we predict the next word $w_{t+1}$ using a two layer bidirectional-LSTM model \cite{lstmHochreiter}. The model is trained on the simple language corpus data as described in \refsec{sec:megdataset} with a cross-entropy loss. We evaluate our model on 10\% held out text data. The perplexity for the Bi-directional Language model is 9.97 on test data (the low perplexity value is due to the simple train and test dataset).
    
    \item \textbf{Multi-task Model}: Motivated by the brain's multitask capability, we build a model to predict next word and POS tag information. The multitask model is a simple two layer bidirectional LSTM model with separate linear layers predicting each of the tasks given the output of the last LSTM layer (\reffig{fig:Multitask_archi}). The model is trained on the simple sentence corpus data as described in \refsec{sec:megdataset} with a cross-entropy loss. The model's accuracy is 96.9\% on the POS-tag prediction task and has perplexity of %9.085 
    	9.09 on the 10\% test data. The high accuracy and low perplexity are due to the simple nature of our language dataset.
    
    \item \textbf{ELMO}~\cite{elmo2018}: 
    ELMo is a recent state-of-the-art deep contextualized word representation method which models a word's features as internal states of a deep bidirectional language model (biLM) pre-trained on a large text corpus. The contextualized word vectors are able to capture interesting word characteristics like polysemy. ELMO has been shown to improve performance across multiple tasks, such sentiment analysis and question answering.

    \item \textbf{BERT}~\cite{bert2018}: BERT uses a novel technique called Masked Language Model (MLM). MLM randomly masks some tokens inputs and then predicts them. Unlike previous models, this technique can use both left and right context to predict the masked token. The training also predicts the next sentence. The embedding in this model consists of 3 components: token embedding, sentence embedding and transformer positional embedding. Due to the presence of sentence embeddings, we observe an interesting performance of the embedding layer in our experiments. 
\end{itemize}

\section{Experiments and Results}
\label{sec:exp_res}

With human brain as the reference language processing engine, we investigate the relationship between deep neural network representation and brain activity recorded while processing the same sentence. For this task, we perform experiments at both the macro and micro sentence context level. The \textbf{macro-context} experiments evaluate the overall performance of deep neural networks in predicting brain data for input words (all words, nouns, verbs etc.). The \textbf{micro-context} experiments, by contrast, focus on evaluating the performance of deep neural network representations in detecting minor changes in sentence context prior to the token being processed.

%\reminder{PPT: need to define macro and micro contexts first before going into experiments} \reply{SJ: added 2 lines summarising the macro and micro context experiments}

%\reminder{PPT: please flesh out training and testing a little more, too many details are outsourced to the previous papers. The current paper needs to be self-contained, especially w.r.t. relevant details.}

\textbf{Regression task}: Similar to previous research  \cite{2008Mitchell,WehbeEMNLP14}, we use a classification task to align model representations with brain data. MEG data (\refsec{sec:megdataset}) is used for these experiments. The task classifies between a candidate word and the true word a subject is reading at the time of brain activity recording. The classifier uses an intermediate regression step to predict the MEG activity from deep neural network representation for the true and the candidate word. The classifier then chooses the word with least Euclidean distance between the predicted and the true brain activity. A correct classification suggests that the deep neural network representation captures important information to differentiate between brain activity at words in different contexts. Detailed steps of this process are described as follows. 

    \textbf{Regression training:} We perform regression from the neural-network representation (for each layer) to the brain activity for the same input words in context. We normalized, preprocessed and trained on the MEG data as  described by \cite{WehbeEMNLP14} (Section 2.3.2). We average the signal from every sensor (total 306) over 100ms non-overlapping windows, yielding a $306 \times 5$ sized MEG data for each word. To train the regression model, we take the training portion of the data in each fold, $(X,Y)$, in the tuple $(x_i,y_i)$, $x_i$ is the layer representation for an input word $i$ in a neural network model, and $y_i$ is the corresponding MEG recording of size 1530 (flattened 306*5). The Ridge regression model (f) \cite{scikit-learn} is learned with generalized cross-validation to select $\lambda$ parameter \cite{gcvGolub}. Ridge regression model's $\alpha$ parameter is selected from range $[0.1, \ldots,100 ,1000 ]$. The trained regression model is used to estimate MEG activity from the stimulus features, i.e., $\hat{y_i} = f(x_i)$.
    
\textbf{Regression testing}:
    The trained regression model is used to predict $\hat{y_i}$ for each word stimulus ($x_i$) in the test fold during cross-validation. We perform a pair-wise test for the classification accuracy (Acc) \cite{2008Mitchell}. The chance accuracy of this measure is 0.5. We use Euclidean distance ($\text{E}_{dist}$) as given in \eqref{eq:1} for the measure.
    
%     \begin{equation}
% \begin{aligned}
%         Score = 
%         \begin{cases}
%     1,& \text{if } dist(f(x1),y1)+dist(f(x2),y2) 
    
%     \geq  dist(f(x1),y2)+dist(f(x2),y1)\\
%     0,              & \text{otherwise}
%         \end{cases}
%         \end{aligned}
% \end{equation}

\begin{equation} \label{eq:1}
  \mathrm{Acc} = \left\{\begin{alignedat}{2}
    & 1, && \ \text{if } \text{E}_{dist}(f(x_i),y_i) + \text{E}_{dist}(f(x_j),y_j) \\
    & && \leq  \text{E}_{dist}(f(x_i),y_j) + \text{E}_{dist}(f(x_j),y_i) \\
    & 0, &&\quad \text{otherwise}
  \end{alignedat}\right.
\end{equation}
  
\begin{figure*}[t]
\centering
  \setlength{\textfloatsep}{10.0pt}
  \includegraphics[scale=0.45]{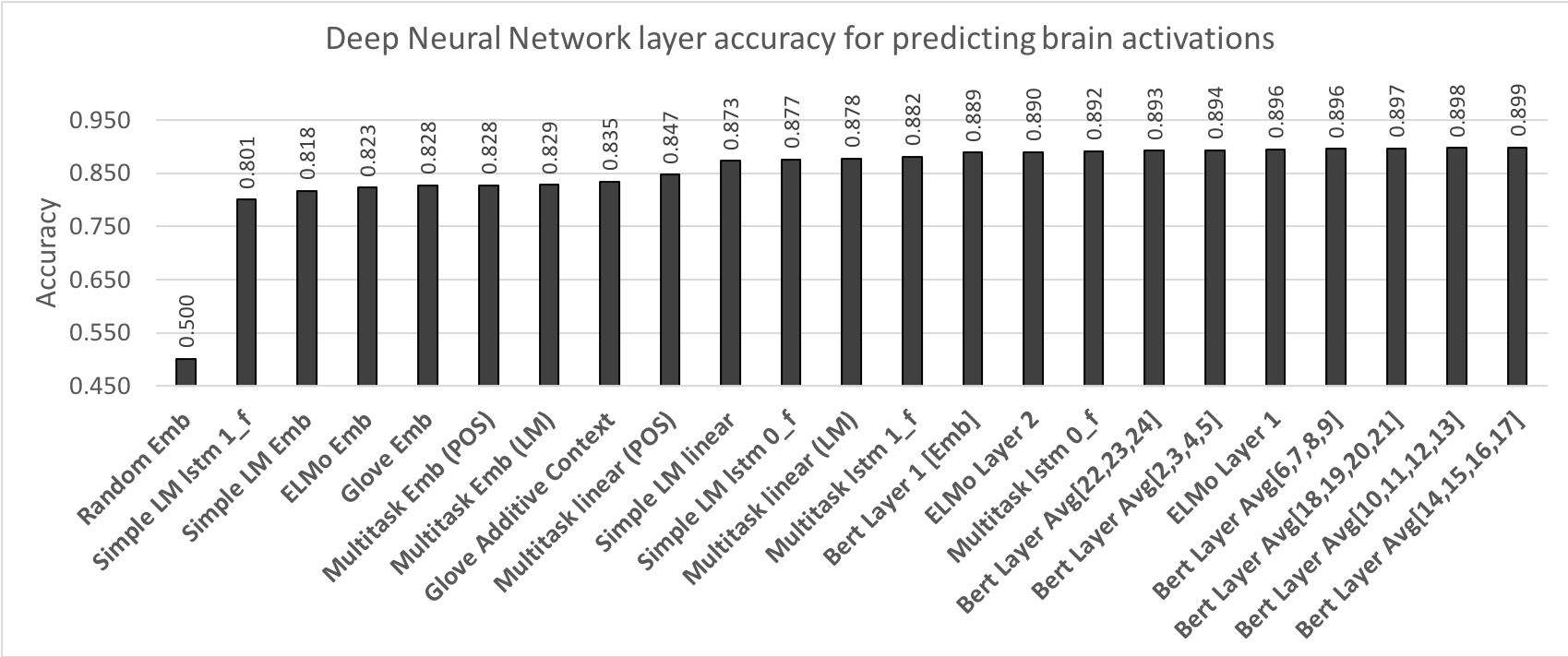}
  \caption{\label{fig:Encoding_results} Pairwise classification accuracy of brain activity data predicted from various model layer representations. We average 4 consecutive layers of BERT into one value. We find that BERT and ELMO model layers perform the best. The middle layers of most models and BERT, in particular, are good at predicting brain activity. Read `\_f' as forward layer and `Emb' as the embedding layer.}
\end{figure*}

\begin{figure*}[t]
\centering
  \setlength{\textfloatsep}{10.0pt}
  \includegraphics[scale=0.45]{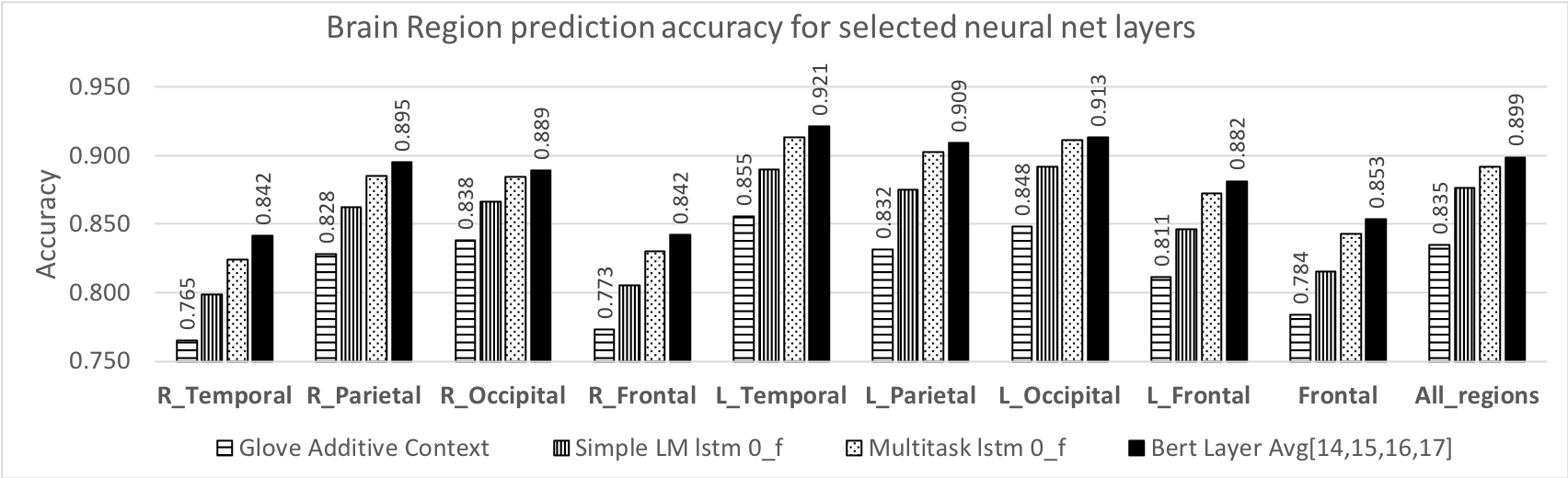}
  \caption{\label{fig:Encoding_result_brain} Pairwise accuracy of various brain regions from some selected deep neural network model layers. The left part of the brain which is considered central to language understanding is predicted with higher accuracy, especially left temporal region (L = left, R = right).}
\end{figure*}

\subsection{Macro-context Experiments}

%A number of deep neural network models have been proposed to process natural language. We compare the representations learnt by these models and their individual layers to gain insight about sentence representation by aligning them with human brain activity (reference data). 
The macro-context experiments aggregate classification performance of each model's layer on the entire stimuli set. We also evaluate on smaller sets such as only the nouns, verbs, passive sentence words, active sentence words, etc. The macro experiments help us to compare all the models on a large stimuli set. In summary, we observe the following: (1) the intermediate layers of state-of-the-art deep neural network models are most predictive of brain activity (\citet{NIPS2018_lstm_fmri} also observe this on a 3 layer LSTM language model), (2) in-context representations are better at predicting brain activity than out-of-context representations (embeddings), and (3) Temporal lobe is predicted with highest accuracy from deep neural network representations. 

\textbf{Detailed Observations}: The results of pairwise classification tests for various models are presented in \reffig{fig:Encoding_results}. All the results reported in this section are for \passacttwo{}. From the figure, we observe that BERT and ELMo outperform the simple models in predicting brain activity data. In the neural network language models, the middle layers perform better at predicting brain activity than the shallower or deeper layers. This could be due to the fact that the shallower layers represent low-level features and the deeper layers represent more task-oriented features. We tested this hypothesis by examining the performance scores at each lobe of the brain. For each area, we tested the left and right hemispheres independently and compared these performances with the bilateral frontal lobe as well as the activity across all regions. In particular, we examined the primary visual areas  (left and right occipital lobe), speech and language processing areas (left temporal) and verbal memory (right temporal), sensory perception (left parietal) and  integration (right parietal), language related movements (left frontal) and non-verbal functioning (right frontal). The frontal lobe was tested bilaterally as it is associated with higher level processing such as problem solving, language processing, memory, judgement, and social behavior.

% \begin{itemize}
%     \setlength\itemsep{-.2em}
%     \item L\_Occipital: Visual processing center.
%     \item R\_Occipital: Visual processing center.
%     \item L\_Temporal: Speech and language processing.
%     \item R\_Temporal: Right temporal lobe plays a role in verbal memory \cite{rt_func}.
%     \item L\_Parietal: Sensation and perception;
%     \item R\_Parietal: Integrating sensory input.
%     \item L\_Frontal:  Language related movement.
%     \item R\_Frontal: Non-verbal abilities.
%     \item Frontal:  motor function, problem solving, spontaneity, memory, language, initiation, judgement, impulse control, and social behavior.
% \end{itemize}

From our results, we observe that lower layers such as BERT layer 5 have very high accuracy for right occipital and left occipital lobe associated with low-level visual processing task. In contrast, higher layers such as linear layers in the Multitask Model and in Language Model have the highest accuracy in the left temporal region of the brain. \reffig{fig:Encoding_result_brain} shows the pairwise classification accuracy for a given brain region for best layers from each model. The accuracy is highest in left temporal region, responsible for syntactic and semantic processing of language. These results establish correspondences between representations learned by deep neural methods and those in the brain. %These results show that the current deep learning language models understand sentence context in ways similar to the human brain. 
Further experiments are needed to improve our understanding of this relationship. %\reminder{PPT: we may downplay the previous claim, and just say that this **may** be one of the causes which requires further work} \reply{SJ: added one line in the end that futher experiments are needed }

We performed additional experiments to predict on a restricted stimuli set. In each of these experiments, a subset of stimuli, for example active sentences, passive sentences, noun, and verb stimuli were used in classification training and testing. Detailed results for this experiment are documented in the appendix section (\reffig{fig:macro-context}). From the results, we observe that active sentences are predicted better (best accuracy = 0.93) than passive sentences (best accuracy = 0.87). This might be attributed to the nature of training datasets for deep neural networks, as active sentences are dominant in the training data of most of the pre-trained models. We also observe that for passive sentences, our simple multitask model (trained using about 250K active and passive sentences) has a lower performance gap between active and passive sentence as compared to ELMO and BERT models. This may be due to a more balanced active and passive sentence used to train the multitask model. Noun stimuli are predicted with the highest accuracy of 0.81, while the accuracy for verbs is 0.65. Both  Multitask and ELMo models dominate verb prediction  results, while BERT lags in this category. Further experiments should be done to compare the ability of Transformer \cite{NIPS:Attention} versus Recurrent Neural Network based models to represent verbs.

\subsection{Micro-context Experiments}
\label{sec:micro_test}

%\reminder{PPT: Please fix the informal language below and explain more the training part.} \reply{Some informal sentences removed.}

%We test, if our models correctly retain context from each word of the sentence in the past, using fine-grained context sensitive tests. 
In these micro-context experiments, we evaluate if our models are able to retain information from words in the sentence prior to the word being processed. For such context sensitivity tests, we only use the first repetition of the sentence shown to human subjects. This helps to ensure that the sentence has not been memorized by the subjects, which might affect the context sensitivity tests. 

\begin{figure*}[t]
\centering
  \setlength{\textfloatsep}{10.0pt}
  \includegraphics[scale=0.5]{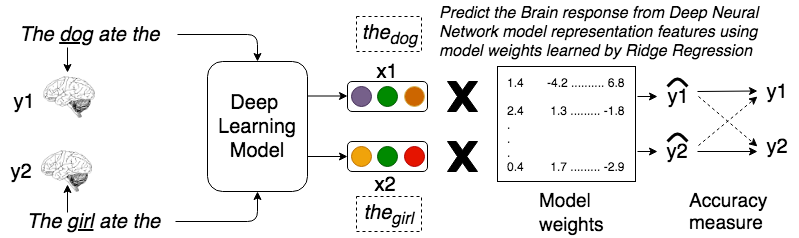}
  \caption{\label{fig:micro_context_test} Experimental setup for micro-context tests. Given two sentences with similar words except one in the past (underlined), the test evaluates if the deep neural network model representation contains sufficient information to tell the two words apart. Please see \refsec{sec:micro_test} for more details.}
\end{figure*}

\textbf{Training:} The micro-context experiment setup  is illustrated in \reffig{fig:micro_context_test}. 
%\reminder{PPT: fold and tuple is confusing in next sentence. Do we really need the fold variables? Also all models are assumed to be neural, but that is not always the case as one of the baselines is random} 
To train the regression model, each training instance corresponding to a word has the form $(x_i,y_i)$, % we take the training portion of the data in each fold, $(X,Y)$, where the tuple $(x_i,y_i)$ is a word. 
where $x_i$ is the layer representation for an input word $i$ in a neural network model, and $y_i$ is the corresponding MEG brain recording data of size 1530 (flattened $306 \times 5$). During testing, we restrict the pairwise tests to word pairs ($x_i,x_j$) which satisfy some conditions. For example in noun context sensitivity test, the pair of words should be such that, they appear in a sentence with the same words except the noun. We describe these candidate word test pairs, in detail, in the following sections.  

In each of the following sensitivity tests, we perform a pair-wise accuracy test among the same candidate word (bold items) from sentences which are identical except for one word (underlined items). We vary the non-identical word type (noun, verb, adjective, determiner) among the two sentences to test the contribution of each of these word types to the context representation further in sentence. This test helps us understand what parts of the context are retained or forgotten by the neural network model representation. Detailed results of each test are included in the appendix section (\reffig{fig:micro-context}). Please note that the part of BERT word embedding is the sentence embedding, therefore the BERT embedding performs better than 0.5, unlike other embeddings.

\subsubsection{Noun sensitivity}
\label{sec:noun_sensitivity}

\begin{center}
 ``\textit{The \underline{dog} ate \textbf{the}}" \quad vs. \quad
 ``\textit{The \underline{girl} ate \textbf{the}}"
\end{center}

%\reminder{PPT: the accuracy numbers for sensitivity experiments are based on what, which table?} \reply{SJ: mentioned in the paragraph above, Appendix section, Figure  10}

For the \passacttwo{}, we observe that simple GloVe additive model (classification accuracy = 0.52) loses information about the noun while it is retained by most layers of other models like BERT (accuracy = 0.92), ELMo (accuracy = 0.91). Higher level layers, such as linear layer for POS-tag prediction (accuracy = 0.65), also perform poorly. This seems obvious due to the task it solves which focuses on POS-tag property at the word `the' rather than the previous context. In summary, we observe that the language model context preserves noun information well. 

\subsubsection{Verb sensitivity}
\label{sec:verb_sensitivity}

\begin{center}
``\textit{The dog \underline{saw} \textbf{the}}"\quad vs. \quad
 ``\textit{The dog \underline{ate} \textbf{the}}"
\end{center}
 
For the \passacttwo{}, we observe that similar to noun sensitivity, most language model layers (accuracy = 0.92), except for simple GloVe Additive model, preserve the verb memory. By design, the GloVe Additive model retains little context from the past words, and therefore the result verifies the experiment setup.

\subsubsection{First determiner sensitivity}

\begin{center}
``\textit{\underline{A} \textbf{dog}}" \quad vs. \quad  ``\textit{\underline{The} \textbf{dog}}"
\end{center}

For the \krnstwo{}, we observe that determiner information is retained well by most layers. However, the shallow layers retain information better than the deeper layers. For example, BERT layer 3 (accuracy = 0.82), Multitask lstm 0\_backward (accuracy = 0.82), BERT Layer 18/19 (accuracy 0.78). Since the earlier layers have a higher correlation with shallow feature processing, the determiner information may be useful for the early features in neural network representation.

\subsubsection{Adjective sensitivity}

\begin{center}
``\textit{The \underline{happy} \textbf{child}}"\quad vs. \quad ``\textit{The \textbf{child}"}
\end{center}

For the \krnsfive{}, we observe that middle layers of most models (BERT, Multitask) retain the adjective information well. However, surprisingly simple multitask model (lstm 1\_forward layer accuracy = 0.89) retains adjective information better than BERT model (layer 7 accuracy = 0.84). This  could be due to the importance of adjective in context for POS tag prediction. This result encourages the design of language models with diverse cost functions based on the kind of sentence context information that needs to be preserved in the final task. 

\subsubsection{Visualisation}

We visualise the average agreement of model predicted brain activity (from BERT layer 18) and true brain activity for candidate stimuli in micro-sensitivity tests. Please note that the micro-sensitivity tests predict brain activity for stimuli with almost similar past context except one word, this makes the task harder. We preprocess the brain activity values to be +1 for all positive values and -1 for all negative values. The predicted brain activity ($y^{'}$) and the true brain activity ($y$) are then compared to form an agreement activity ($y^{''}$), resulting in a zero value for all locations where the sign predicted was incorrect. We average these agreement activities ($y^{''}$) for all test examples in a cross-validation fold to form a single activity image ($Y^{''}$). \reffig{fig:noun_delta_brain} shows $Y^{''}$ for the word `\textit{the}' in noun-sensitivity tests \refsec{sec:noun_sensitivity} (additional results are in the appendix section). %\reffig{fig:verb_delta_brain} shows Y'' for the word 'the' in verb-sensitivity tests \refsec{\label{sec:verb_sensivity}}. 
We observe that our model prediction direction agrees with brain prediction direction in most of the brain regions. This shows that our neural network layer representation can preserve information from earlier words in the sentence. 

%\reminder{PPT: we may summarize all sensitivity observations in a table}\reply{SJ: not sure, what parts can be tabulated, please explain further.}

\begin{figure*}[t]
\centering
  \setlength{\textfloatsep}{10.0pt}
  \includegraphics[scale=0.45]{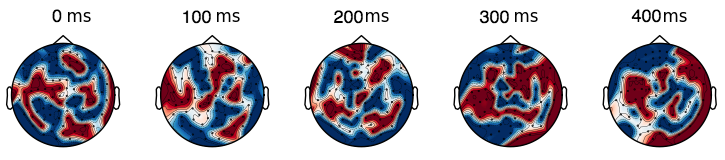}
  \caption{\label{fig:noun_delta_brain} Average sign agreement activity for noun sensitivity stimuli `\textit{the}'. The red and blue colored areas are the +ive and -ive signed brain region agreement respectively, while the white colored region displays brain regions with prediction error. We observe that in most regions of the brain, the predicted and true activity agree on the activity sign, thereby providing evidence that deep learning representations can capture useful information about language processing consistent with the brain recording. %\reminder{PPT: check claim in last sentence}
}

\end{figure*}

\subsection{Semi-supervised training using synthesized brain activity}
\label{sec:synth_brain}

\begin{figure*}[t]
\begin{subfigure}{.5\textwidth}
  \centering
  \includegraphics[scale=0.43]{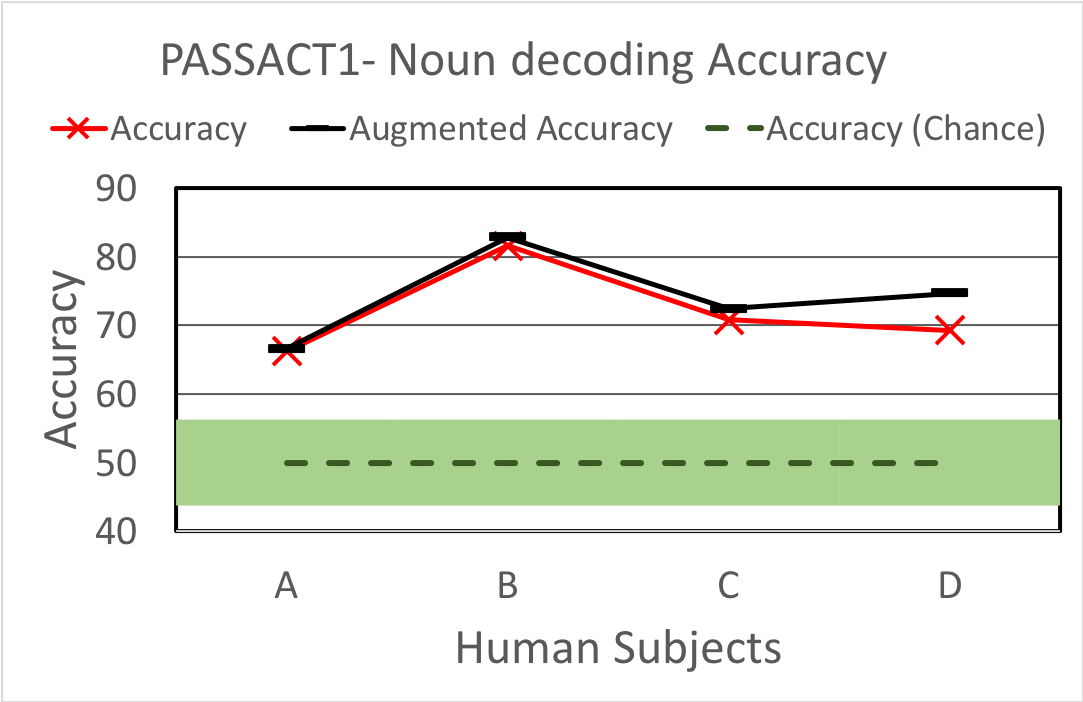}
  \caption{Noun prediction results}
 \end{subfigure}
\begin{subfigure}{.5\textwidth}
  \centering
   \includegraphics[scale=0.43]{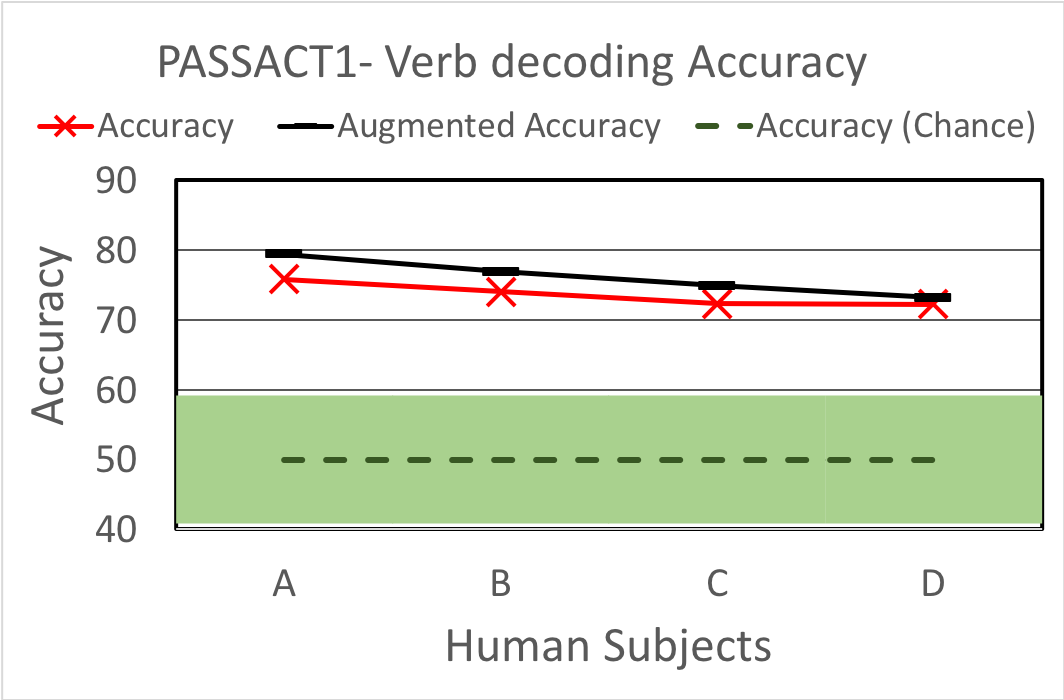}
  \caption{Verb prediction results}
\end{subfigure}
\caption{\label{fig:decoding_results}Accuracy with and without synthetically generated MEG brain data on two stimuli prediction tasks: (a) Nouns (left) and (b) Verbs (right). We trained two models -- one  using true MEG brain recording and the other using both true and synthetically generated MEG brain data (Augmented data model). We observe that the augmented data model results in accuracy improvement on both tasks, on average \textbf{2.1\%} per subject for noun prediction and \textbf{2.4\%} for verb. Accuracy (chance) is the random permutation test accuracy, with the green shaded area representing standard deviation. Please see Section \ref{sec:synth_brain} for details.%Plotted values are the accuracy results on stimuli prediction tasks. We train one model using true MEG brain recording and the second model using both true MEG brain recording and synthetically generated MEG brain data. Testing on the same data (test fold of CV) the augmented data training gives better stimuli prediction accuracy. Results improve by an average \textbf{2.1\%} per subject for noun prediction and \textbf{2.4\%} for verb. Accuracy (chance) is the random permutation test accuracy, the green shaded area shows it's standard deviation. 
%\reminder{PPT: in the legend you may say Accuracy (Brain Only)}}\reply{SJ: space issues in plot:(, mentioned it in caption}
}
\end{figure*}

%Results from the previous section show that deep neural network models can be used to predict difference in brain activity that are due to earlier difference in sentence due to noun, verb, determiner and adjectives in the sentence. 
In this section, we consider the question of whether  previously trained linear regression model (X1),  which predicts brain activity for a given sentence, %word-by-word 
can be used to produce useful synthetic brain data (i.e.,  sentence-brain activity pairs). Constraints like high cost of MEG recording and physical limits on an individual subject during data collection, favor such synthetic data generation. We evaluate effectiveness of this synthetically generated brain data for data augmentation in the stimulus prediction task \cite{2008Mitchell}. Specifically, we train a decoding model (X2) to predict brain activity during a stimulus reading based on GloVe vectors for nouns. We consider two approaches.  %in two ways. One way is to use the same training brain activity data as in the previous sections. The second way is to again use this same data, but augment it with synthetic training examples produced by the model (X1). 
In the first approach, the same brain activity data as in previous sections was used. In the second approach, the real brain activity data is augmented with the synthetic activities generated by the regression model (X1).

% Our current results favor the neural network models for the right understanding of context. In the tests we generate brain activation data from the neural network contexts and test these generated brain activation data with respect to true brain activation data. Results in the past sections give us confidence in the ability of these models to predict the correct brain activity. Based on these findings we investigate the use of neural network contexts to generate brain activation data. Brain activation data collection is limited by the human subject's capacity and the cost associated with the imaging device. The generated brain activation data, which is almost zero cost (computer algorithms) can be used to augment the currently available brain activation data in statistical analysis and pattern recognition.\reminder{Shar: semi-supervised?}

In our experiment, we generate new sentences using the same vocabulary as the original sentences in the  \passacttwo{}. Details of the original 32 sentences (\refsec{supp:passact2}) along with the 160 generated sentences (\refsec{supp:passact2_aug}) are given in the appendix section. We process the 160 generated sentences with BERT layer 18 to get word stimulus features in context. The encoding model (X1) was trained using the \passacttwo{}. 
Please note that BERT layer 18 was chosen based on the high accuracy results on macro-context tests, therefore the layer aligned well with the whole brain activity. The choice of representation (deep neural network layer) to encode brain activity should be done carefully, as each representation may be good at encoding different parts of brain. A good criteria for representation selection requires further research.

%\reminder{PPT: please rephrase the following para.}\reply{SJ: Rephrased some parts of it. Hopefully it flows better now}

To demonstrate the efficacy of the synthetic dataset, we present the accuracy in predicting noun (or verb) stimuli from observed MEG activity with and without the additional synthetic MEG data. With linear ridge regression model (X2), a GloVe \cite{pennington2014glove} feature to brain-activity prediction models were trained to predict the MEG activity when a word is observed . To test the model performance, we calculate the accuracy of the predicted brain activity given the true brain activity during a word processing (Equation \ref{eq:1}). All the experiments use 4-fold cross-validation. \reffig{fig:decoding_results} shows the increase in the noun/verb prediction accuracy with additional synthetically generated data. The statistical significance is calculated over 400 random label permutation tests. 

To summarize, these results show the utility of using  previously trained regressor model to produce synthetic training data to improve accuracy on additional tasks. Given the high cost of collecting MEG recordings from human subjects and their individual capacity to complete the task, this data augmentation approach may provide an effective  alternative in many settings.

%\reminder{PPT: instead of X1 and X2, just use proper names. They don't add much value. Also, give motivation for high cost at the start of Sec 4.3 as well.} 
%\reply{SJ: Added the motivation at the start of section 4.3. The X1 and X2 models were added to distinguish the between the encoding and decoding models, one of the reviewers had some confusion here.}

\section{Related Work}

% General Matching of brain activity in language processing (words, sentences, reading, ERP)
Usage of machine learning models in neuroscience has been gaining popularity. Methods in this field use features of words and contexts to predict brain activity using various techniques \cite{AgrawalSMG14}. Previous research have used functional magnetic resonance imaging (FMRI) \cite{fmri_ref} and Magnetoencephalography (MEG) \cite{meg_ref} to record brain activity.
%MEG is especially ideal for understanding the fast dynamics of language processing due to it's time resolution. 
% Matching computer program states to brain activity (CNNs+Vision)
%Some paper in current literature have used machine learning and deep learning in particular to form model hypothesis and study parallels between these models and brain to form theories.
%Paper by \citet{mante_nature} studies the prefrontal cortex in rhesus monkeys. 
Prefrontal cortex in rhesus monkeys was studied in \citet{mante_nature}. They showed that an appropriately trained recurrent neural network model reproduces key physiological observations and suggests a new mechanism of input selection and integration. 
\citet{BARAK20171} argues that RNNs with reverse engineering can provide a framework for modeling in neuroscience, potentially serving as a powerful hypothesis generation tool.
%\cite{vis_dnn_Kriegeskorte}
%In \cite{NIPS2018_rnnconv} authors show that the proposed ConvRNNs, task-driven convolutional recurrent neural networks provide the best models of encoding dynamics in the primate visual system.
%, suggesting a role for the brain’s recurrent connections in performing difficult visual behaviors.
% Brain Context Understanding (Leila's EMNLP 2014, NIPS 2018)
%NLP papers such as 
Prior research by \citet{2008Mitchell}, \citet{WehbeEMNLP14},  \citet{NIPS2018_lstm_fmri}, \citet{hale_acl18}, \citet{Pereira:nature},  and \citet{sun2019} have established a general correspondence between a computational model and brain's response to naturalistic language. We follow these prior research in our analysis work and extend the results by doing a fine-grained analysis of the sentence context. Additionally, we also use  deep neural network representations to generate synthetic brain data for extrinsic experiments.

\section{Conclusion}

In this paper, we study the relationship between sentence representations learned by deep neural network models and those encoded by the brain. We encode simple sentences using multiple deep networks, such as ELMo, BERT, etc. We make use of MEG brain imaging data as  reference. Representations learned by BERT are the most effective in predicting brain activity. In particular, most models are able to predict activity in the left temporal region of the brain with high accuracy. This brain region is also known to be responsible for processing syntax and semantics for language understanding. \reminder{PPT: next sent is not clear. Hmm, still not clear}\reply{SJ: updated} To the best of our knowledge, this is the first work showing that the MEG data, when reading a word in a sentence, can be used to distinguish earlier words in the sentence. Encouraged by these findings, we use deep networks to generate synthetic brain data to show that it helps in improving accuracy in a subsequent stimulus decoding task. Such data augmentation approach is very promising as actual brain data collection in large quantities from human subjects is an expensive and labor-intensive process. We are hopeful that the ideas explored in the paper will promote further research in understanding relationships between representations learned by deep models and the brain during language processing tasks. 

% In the paper, we use multiple language models to predict brain activity. We find that the deep neural network representations predicts brain activity well. In particular most models predict the left temporal region of the brain with high accuracy. This brain region is known to be responsible for processing syntax and semantics for language understanding. 
%To our knowledge this is the first work showing that the MEG brain activation when reading a word in a sentence can be used to distinguish earlier words in the sentence. Furthermore, as shown in \reffig{fig:noun_delta_brain} the the difference in brain activity are correctly predicted in terms of it's sign by the trained model from deep neural network representation in most brain regions. 
%Encouraged by these findings we use deep neural network representation vector to generate artificial brain data for additional sentences. The generated brain data is used along with existing data to improve noun and verb brain activity prediction accuracy. We find these results encouraging and hope to use deep learning methods to help in brain understanding experiments further. 
\section{Acknowledgments}
This work was supported by The Government of India (MHRD) scholarship and BrainHub CMU-IISc Fellowship awarded to Sharmistha Jat. We thank Dan Howarth and Erika Laing for help with MEG data preprocessing.
% %-------------------------
\clearpage
\bibliography{acl2019}
\bibliographystyle{styles/acl_natbib}
%-------------------------
\clearpage
\appendix
\section{Appendices}
%\label{sec:appendix}
%Appendices are material that can be read, and include lemmas, formulas, proofs, and tables that are not critical to the reading and understanding of the paper. 

%\section{Supplemental Material}
%\label{sec:supplemental}
\subsection{Dataset details}
\label{supp:dataset3}
Following are the sentences used in the paper for experiments described in \refsec{sec:exp_res}. We list down the sentences in \passacttwo{} and the generated sentences in the sections \refsec{supp:passact2} and \refsec{supp:passact2_aug} respectively. The two datasets are disjoint in terms of the sentences they contain, but are built using the same vocabulary. Datasets \krnstwo{} and \krnsfive{} are detailed in subsections \ref{supp:krns2} and \ref{supp:krns5} respectively. 

\subsubsection{\passacttwo{} sentences}
\label{supp:passact2}
the boy was liked by the girl \\
the girl was watched by the man  \\   
the man was despised by the woman  \\   
the woman was encouraged by the boy  \\   
the girl was liked by the woman   \\  
the man was despised by the boy   \\  
the girl was liked by the boy    \\ 
the boy was watched by the woman  \\   
the man was encouraged by the girl \\    
the woman was despised by the man  \\   
the woman was watched by the boy  \\   
the girl was encouraged by the woman  \\   
the man was despised by the girl \\    
the boy was liked by the man   \\  
the boy was watched by the girl  \\   
the woman was encouraged by the man   \\  
the man despised the woman   \\  
the girl encouraged the man  \\   
the man liked the boy     \\
the girl despised the man   \\  
the woman encouraged the girl  \\   
the boy watched the woman  \\   
the man watched the girl   \\  
the girl liked the boy   \\  
the woman despised the man  \\   
the boy encouraged the woman  \\   
the woman liked the girl  \\   
the boy despised the man  \\   
the man encouraged the woman  \\   
the girl watched the boy   \\  
the woman watched the boy  \\   
the boy liked the girl  \\
\subsubsection{\passacttwo{} artificially generated sentences}
\label{supp:passact2_aug}
the girl was despised by the man \\
the man despised the girl \\
the man was liked by the girl \\
the girl was liked by the man \\
the girl liked the man \\
the man liked the girl \\
the girl was encouraged by the man \\
the man encouraged the girl \\
the man was watched by the girl \\
the girl watched the man \\
the boy was despised by the man \\
the man despised the boy \\
the man was liked by the boy \\
the boy liked the man \\
the man was encouraged by the boy \\
the boy was encouraged by the man \\
the boy encouraged the man \\
the man encouraged the boy \\
the man was watched by the boy \\
the boy was watched by the man \\
the boy watched the man \\
the man watched the boy \\
the man was despised by the women \\
the women was despised by the man \\
the women despised the man \\
the man despised the women \\
the man was liked by the women \\
the women was liked by the man \\
the women liked the man \\
the man liked the women \\
the man was encouraged by the women \\
the women was encouraged by the man \\
the women encouraged the man \\
the man encouraged the women \\
the man was watched by the women \\
the women was watched by the man \\
the women watched the man \\
the man watched the women \\
the girl was despised by the man \\
the man despised the girl \\
the girl was liked by the man \\
the man was liked by the girl \\
the man liked the girl \\
the girl liked the man \\
the girl was encouraged by the man \\
the man encouraged the girl \\
the man was watched by the girl \\
the girl watched the man \\
the girl was despised by the boy \\
the boy was despised by the girl \\
the boy despised the girl \\
the girl despised the boy \\
the girl was encouraged by the boy \\
the boy was encouraged by the girl \\
the boy encouraged the girl \\
the girl encouraged the boy \\
the girl was watched by the boy \\
the boy watched the girl \\
the girl was despised by the women \\
the women was despised by the girl \\
the women despised the girl \\
the girl despised the women \\
the girl was liked by the women \\
the women was liked by the girl \\
the women liked the girl \\
the girl liked the women \\
the girl was encouraged by the women \\
the women was encouraged by the girl \\
the women encouraged the girl \\
the girl encouraged the women \\
the girl was watched by the women \\
the women was watched by the girl \\
the women watched the girl \\
the girl watched the women \\
the boy was despised by the man \\
the man despised the boy \\
the man was liked by the boy \\
the boy liked the man \\
the boy was encouraged by the man \\
the man was encouraged by the boy \\
the man encouraged the boy \\
the boy encouraged the man \\
the boy was watched by the man \\
the man was watched by the boy \\
the man watched the boy \\
the boy watched the man \\
the boy was despised by the girl \\
the girl was despised by the boy \\
the girl despised the boy \\
the boy despised the girl \\
the boy was encouraged by the girl \\
the girl was encouraged by the boy \\
the girl encouraged the boy \\
the boy encouraged the girl \\
the girl was watched by the boy \\
the boy watched the girl \\
the boy was despised by the women \\
the women was despised by the boy \\
the women despised the boy \\
the boy despised the women \\
the boy was liked by the women \\
the women was liked by the boy \\
the women liked the boy \\
the boy liked the women \\
the boy was encouraged by the women \\ 
the women was encouraged by the boy \\
the women encouraged the boy \\
the boy encouraged the women \\
the boy was watched by the women \\
the women was watched by the boy \\
the women watched the boy \\
the boy watched the women \\
the women was despised by the man \\
the man was despised by the women \\
the man despised the women \\
the women despised the man \\
the women was liked by the man \\
the man was liked by the women \\
the man liked the women \\
the women liked the man \\
the women was encouraged by the man \\
the man was encouraged by the women \\
the man encouraged the women \\
the women encouraged the man \\
the women was watched by the man \\
the man was watched by the women \\
the man watched the women \\
the women watched the man \\
the women was despised by the girl \\
the girl was despised by the women \\
the girl despised the women \\
the women despised the girl \\
the women was liked by the girl \\
the girl was liked by the women \\
the girl liked the women \\
the women liked the girl \\
the women was encouraged by the girl \\
the girl was encouraged by the women \\
the girl encouraged the women \\
the women encouraged the girl \\
the women was watched by the girl \\
the girl was watched by the women \\
the girl watched the women \\
the women watched the girl \\
the women was despised by the boy \\
the boy was despised by the women \\
the boy despised the women \\
the women despised the boy \\
the women was liked by the boy \\
the boy was liked by the women \\
the boy liked the women \\
the women liked the boy \\
the women was encouraged by the boy \\
the boy was encouraged by the women \\
the boy encouraged the women \\
the women encouraged the boy \\
the women was watched by the boy \\
the boy was watched by the women \\
the boy watched the women \\
the women watched the boy \\

\subsubsection{\krnstwo{} sentences}
\label{supp:krns2}
the monkey inspected the peach\\
a monkey touched a school\\
the school was inspected by the student\\
a peach was touched by a student\\
the peach was inspected by the monkey\\
a school was touched by a monkey\\
a doctor inspected a door\\
the doctor touched the hammer\\
the student found a door\\
a student kicked the hammer\\
the student inspected the school\\
a student touched a peach\\
a monkey found the hammer\\
the monkey kicked a door\\
a dog inspected a hammer\\
the dog touched the door\\
a dog found the peach\\
the dog kicked a school\\
the doctor found a school\\
a doctor kicked the peach\\
a school was kicked by the dog\\
the peach was found by a dog\\
the door was touched by the dog\\
a hammer was inspected by a dog\\
the peach was kicked by a doctor\\
a school was found by the doctor\\
the hammer was touched by the doctor\\
a door was inspected by a doctor\\
the hammer was kicked by a student\\
a door was found by the student\\
the hammer was found by a monkey\\
a door was kicked by the monkey\\

\subsubsection{\krnsfive{} sentences}
\label{supp:krns5}
the teacher broke the small camera\\
the student planned the protest\\
the student walked along the long hall\\
the summer was hot\\
the storm destroyed the theater\\
the storm ended during the morning\\
the duck flew\\
the duck lived at the lake\\
the activist dropped the new cellphone\\
the editor carried the magazine to the meeting\\
the boy threw the baseball over the fence\\
the bicycle blocked the green door\\
the boat crossed the small lake\\
the boy held the football\\
the bird landed on the bridge\\
the bird was red\\
the reporter wrote about the trial\\
the red plane flew through the cloud\\
the red pencil was on the desk\\
the reporter met the angry doctor\\
the reporter interviewed the politician during the debate\\
the tired lawyer visited the island\\
the tired jury left the court\\
the artist found the red ball\\
the artist hiked along the mountain\\
the angry lawyer left the office\\
the army built the small hospital\\
the army marched past the school\\
the artist drew the river\\
the actor gave the football to the team\\
the angry activist broke the chair\\
the cellphone was black\\
the company delivered the computer\\
the priest approached the lonely family\\
the patient put the medicine in the cabinet\\
the pilot was friendly\\
the policeman arrested the angry driver\\
the policeman read the newspaper\\
the politician celebrated at the hotel\\
the trial ended in spring\\
the tree grew in the park\\
the tourist hiked through the forest\\
the activist marched at the trial\\
the tourist ate bread on vacation\\
the vacation was peaceful\\
the dusty feather landed on the highway\\
the accident destroyed the empty lab\\
the horse kicked the fence\\
the happy girl played in the forest\\
the guard slept near the door\\
the guard opened the window\\
the glass was cold\\
the green car crossed the bridge\\
the voter read about the election\\
the wealthy farmer fed the horse\\
the wealthy family celebrated at the party\\
the window was dusty\\
the boy kicked the stone along the street\\
the old farmer ate at the expensive hotel\\
the man saw the fish in the river\\
the man saw the dead mouse\\
the man read the newspaper in church\\
the lonely patient listened to the loud television\\
the girl dropped the shiny dime\\
the couple laughed at dinner\\
the council read the agreement\\
the couple planned the vacation\\
the fish lived in the river\\
the flood damaged the hospital\\
the big horse drank from the lake\\
the corn grew in spring\\
the woman bought medicine at the store\\
the woman helped the sick tourist\\
the woman took the flower from the field\\
the worker fixed the door at the church\\
the businessman slept on the expensive bed\\
the businessman lost the computer at the airport\\
the businessman laughed in the theater\\
the chicken was expensive at the restaurant\\
the lawyer drank coffee\\
the judge met the mayor\\
the judge stayed at the hotel during the vacation\\
the jury listened to the famous businessman\\
the hurricane damaged the boat\\
the journalist interviewed the judge\\
the dog ate the egg\\
the doctor helped the injured policeman\\
the diplomat bought the aggressive dog\\
the council feared the protest\\
the park was empty in winter\\
the parent watched the sick child\\
the cloud blocked the sun\\
the coffee was hot\\
the commander ate chicken at dinner\\
the commander negotiated with the council\\
the commander opened the heavy door\\
the old judge saw the dark cloud\\
the young engineer worked in the office\\
the farmer liked soccer\\
the mob approached the embassy\\
the mob damaged the hotel\\
the minister spoke to the injured patient\\
the minister visited the prison\\
the minister found cash at the airport\\
the minister lost the spiritual magazine\\
the mouse ran into the forest\\
the parent took the cellphone\\
the soldier delivered the medicine during the flood\\
the soldier arrested the injured activist\\
the small boy feared the storm\\
the egg was blue\\
the editor gave cash to the driver\\
the editor damaged the bicycle\\
the expensive camera was in the lab\\
the engineer built the computer\\
the family survived the powerful hurricane\\
the child held the soft feather\\
the clever scientist worked at the lab\\
the author interviewed the scientist after the flood\\
the artist shouted in the hotel\\

\begin{figure*}[tbh]

  \begin{subfigure}{.5\textwidth}
  \centering
  \includegraphics[scale=0.5]{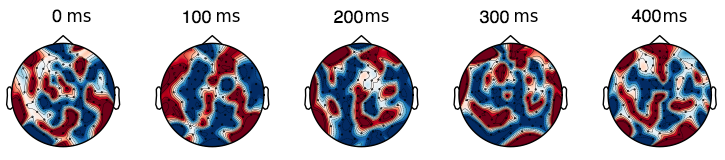}
  \caption{Verb sign agreement image between true and predicted brain activations}
  \end{subfigure}
  
    \begin{subfigure}{.5\textwidth}
    \centering
  \includegraphics[scale=0.5]{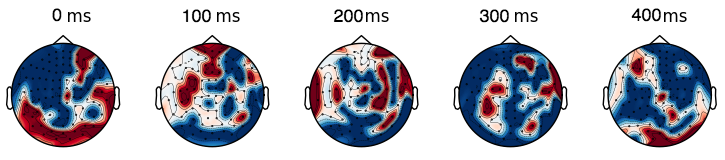}
  \caption{Adjective sign agreement image between true and predicted brain activations}
  \end{subfigure}
  
    \begin{subfigure}{.5\textwidth}
    \centering
  \includegraphics[scale=0.5]{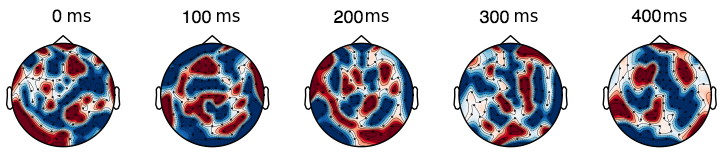}
  \caption{Determiner sign agreement image between true and predicted brain activations}
  \end{subfigure}
  
  \caption{\label{fig:noun_delta_brain} Sign agreement image for verb, determiner and adjective sensitivity test stimuli. The red and blue colored areas are the +ive and -ive signed brain region agreement. While, the white colored region displays brain regions with prediction error. We observe that in most regions of the brain the predicted and true image agree on the activity sign, thereby proving that deep learning representations can capture useful information about language processing.}
\end{figure*}

\begin{figure*}[t]
\centering
  \setlength{\textfloatsep}{10.0pt}
  \includegraphics[scale=0.45]{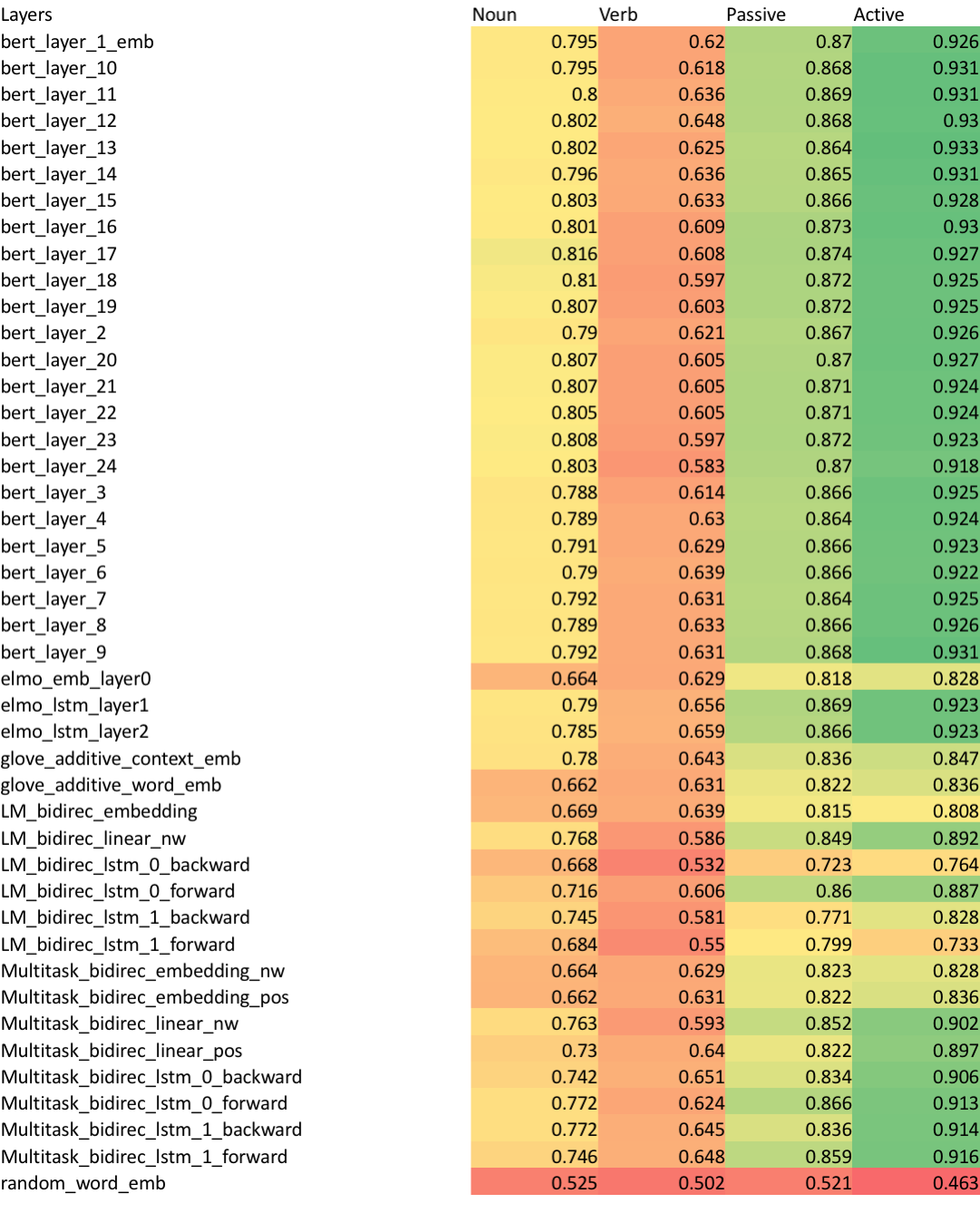}
  \caption{\label{fig:macro-context} Pairwise Accuracy of predicting brain encodings for noun, verb, passive \& active sentences. For each of the category the Ridge regression model is learned and tested on the stimulus subset like only nouns or only passive sentences. The color of a cell represents the value within overall accuracy scale with red indicating small values, yellow intermediate and green high values. We observe that Nouns are predicted better than verbs. And active sentences are predicted better than passive sentences.}
\end{figure*}

\begin{figure*}[t]
\centering
  \setlength{\textfloatsep}{10.0pt}
  \includegraphics[scale=0.45]{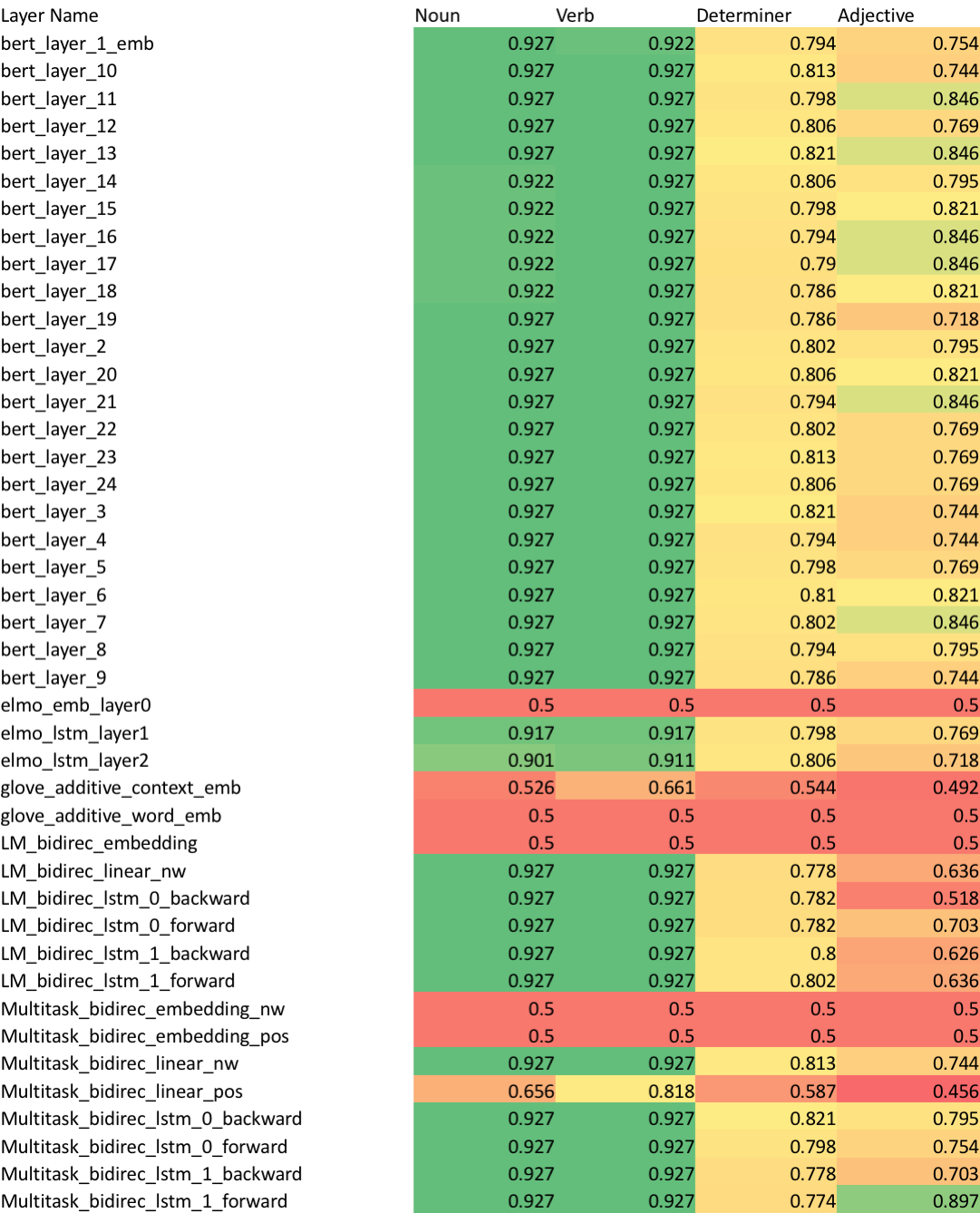}
  \caption{\label{fig:micro-context} Micro-context sensitivity test results for all the layers. The color of a cell represents the value within overall accuracy scale with red indicating small values, yellow intermediate and green high values. We observe that noun and verbs are retained in the context with same accuracy followed by determiner and then adjective.}
\end{figure*}

% \begin{figure*}[t]
% %\setlength{\belowcaptionskip}{-9pt}
% \centering
%   \setlength{\textfloatsep}{10.0pt}
%   \includegraphics[width=16.2cm,height=15.0cm]{diagrams/Architecture/elekta_helmet_sensor}
%   \caption{\label{fig:elekta} Elekta MEG channels corresponding to brain regions.}
% \end{figure*}
\end{document}